\documentclass[letterpaper]{article} 
\usepackage{aaai2026}  
\usepackage{times}  
\usepackage{helvet}  
\usepackage{courier}  
\usepackage[hyphens]{url}  
\usepackage{graphicx} 
\usepackage{natbib}  
\usepackage{caption} 
\usepackage{algorithm}
\usepackage{algorithmic}
\usepackage{amsmath}
\usepackage{amsfonts}
\usepackage{seqsplit}
\usepackage{bbding}
\usepackage{pifont}
\usepackage{booktabs}
\usepackage{multirow}
\usepackage{makecell}
\usepackage{newfloat}
\usepackage{listings}
\DeclareCaptionStyle{ruled}{labelfont=normalfont,labelsep=colon,strut=off} 
\floatstyle{ruled}
\newfloat{listing}{tb}{lst}{}
\floatname{listing}{Listing}
\title{LaF-GRPO: In-Situ Navigation Instruction Generation for the Visually Impaired via GRPO with LLM-as-Follower Reward}
\author {
    Yi Zhao,
    Siqi Wang,
    Jing Li\thanks{Corresponding author.}
}
\affiliations {
    Department of Computing, The Hong Kong Polytechnic University\\
    \{yi-yi-yi.zhao, siqi23.wang\}@connect.polyu.hk, jing-amelia.li@polyu.edu.hk
}

\begin{document}
\maketitle
\begin{abstract}
Navigation instruction generation for visually impaired (VI) individuals (NIG-VI) is critical yet relatively underexplored. This study focuses on generating precise, in-situ, step-by-step navigation instructions that are practically usable for VI users. Specifically, we propose LaF-GRPO (LLM-as-Follower GRPO), where an LLM simulates VI user responses to navigation instructions, thereby providing feedback rewards to guide the post-training of a Vision-Language Model (VLM). This enhances instruction accuracy and usability while reducing costly real-world data collection needs. To address the scarcity of dedicated benchmarks in this field, we introduce NIG4VI, a 27k-sample open-source dataset to facilitate training and evaluation. It comprises diverse navigation scenarios with accurate spatial coordinates, supporting detailed and open-ended in-situ instruction generation. Experiments on NIG4VI demonstrate the effectiveness of LaF-GRPO through quantitative metrics (e.g., Zero-(LaF-GRPO) boosts BLEU 14\%; SFT+(LaF-GRPO) METEOR 0.542 vs. GPT-4o 0.323), and qualitative analysis further confirms that our method yields more intuitive and safer instructions.
\end{abstract}
\begin{links}
\link{Code}{https://github.com/YiyiyiZhao/NIG4VI}
\end{links}
\section{Introduction}
\begin{figure}[t]
  \includegraphics[width=\columnwidth]{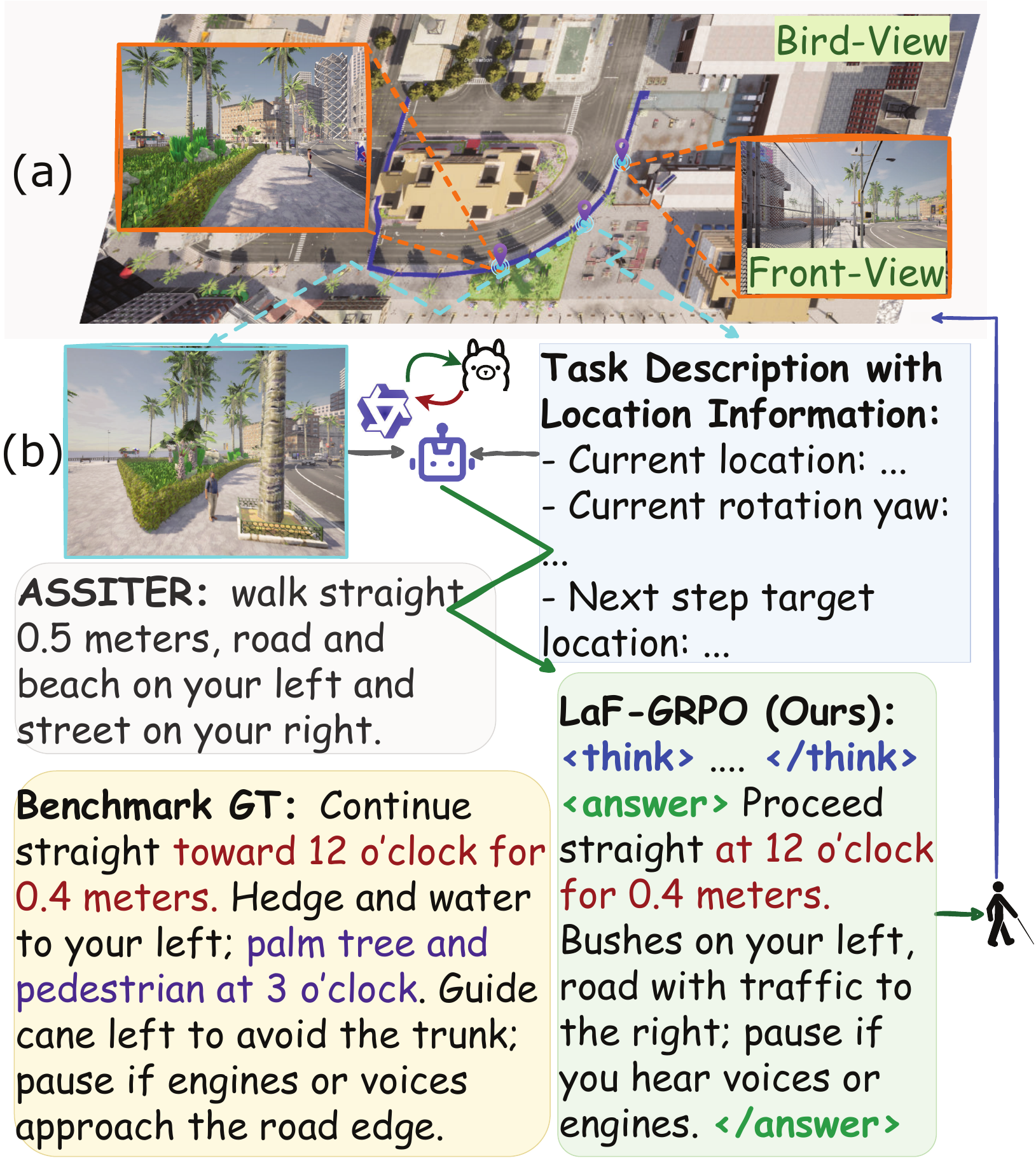}
  \caption{NIG4VI sample. (a) Bird’s-eye map and front views. (b) Instructions generated by ASSISTER~\cite{DBLP:conf/eccv/HuangSZBBO22}, the ground truth, and LaF-GRPO (Ours).}
  \label{fig:task_illustration}
\end{figure}
The Visually Impaired (VI) community, comprising approximately 2.2 billion \cite{WHO2019vision} individuals globally with partial or complete blindness, underscores the significant need for effective assistive technologies. Extensive research in Visually Impaired Assistance (VIA) has emerged to address this need \cite{zhao2024vialmsurveybenchmarkvisually, yuan2025walkvlmaidvisuallyimpairedpeople, gao2025wearable, cai2024navigating}. This paper targets a critical and foundational sub-area: Navigation Instruction Generation for the Visually Impaired (\textbf{NIG-VI}). Unlike Navigation Instruction Generation (NIG) for general embodied agents which produces high-level trajectory plans \cite{DBLP:conf/naacl/DouP22, DBLP:conf/acl/GopinathanMAS24, DBLP:conf/eccv/FanLWY24, DBLP:conf/eccv/KongCWSHYL24}, NIG-VI is fundamentally \textit{human-centered}. As illustrated in Figure~\ref{fig:task_illustration}, the task demands in-situ, step-level instructions that (1) incorporate non-visual cues, (2) ensure precise directional and distance guidance, and (3) adapt to obstacles for safety in map coordinates.

Early attempts, such as ASSISTER \citep{DBLP:conf/eccv/HuangSZBBO22}, laid the initial groundwork in this field but were ultimately constrained by the architectural limitations of BERT-based systems \citep{devlin-etal-2019-bert}. The advent of Vision-Language Models (VLMs) has introduced new opportunities due to their multimodal understanding and generation capabilities. Reinforcement Learning (RL) based post-training methods like GRPO \citep{deepseekai2025deepseekr1incentivizingreasoningcapability} further enhance reasoning abilities, enabling VLMs to align with human preferences, i.e., the human-centered guidance demanded by NIG-VI. However, this alignment process requires collecting large-scale human feedback data for fine-tuning, which can be costly and often fails to incorporate the interactive user feedback essential for achieving human-centered guidance.

To bridge this gap, we propose \textbf{LLM-as-Follower GRPO (LaF-GRPO)}, a novel GRPO-based framework for the NIG-VI task. It features: (1) an LLM that simulates how VI users respond to navigation instructions by interpreting their likely actions, and (2) a VLM post-training procedure for instruction generation guided by an LLM-as-Follower reward. LaF-GRPO mitigates the need for costly trials with VI users while ensuring instruction usability through a \textit{human-in-the-loop} navigation simulation. Furthermore, to address the scarcity of VI navigation benchmarks, we introduce \textbf{NIG4VI}—a comprehensive VI navigation instruction benchmark featuring 27k samples. Fully open-sourced and annotated with granular spatial metadata, NIG4VI enables the generation of detailed, open-ended, in-situ instructions.

Evaluation of LaF-GRPO on the NIG4VI benchmark yields three key findings: 
(1) Qwen2.5-VL models under the \textbf{Zero-(LaF-GRPO)} paradigm significantly outperform the standard zero-shot baseline across multiple metrics. (2) Qwen2.5-VL-7B models trained with \textbf{SFT+(LaF-GRPO)} achieve state-of-the-art performance, reaching a METEOR score of 0.542 and substantially surpassing strong proprietary models such as GPT-4o. (3) Beyond quantitative gains, qualitative analysis shows that LaF-GRPO produces more human-centered instructions, characterized by greater linguistic diversity, more intuitive directional cues, richer environmental details, and essential safety considerations. In summary, our main contributions are:

~$\bullet$ The LaF-GRPO framework, the first to employ GRPO for NIG-VI with a LLM-simulated follower feedback.

~$\bullet$ The NIG4VI benchmark, the first open-source comprehensive dataset with precise multi-modal navigation contexts for robust model evaluation for VI navigation.

~$\bullet$ Extensive empirical studies across VLMs under various paradigms (Zero-shot, Zero-(LaF-GRPO), SFT, and SFT+(LaF-GRPO)), validating the method's effectiveness.
\section{Related Work}
\textbf{VLMs} \citep{DBLP:conf/nips/LiuLWL23a, DBLP:conf/nips/Dai0LTZW0FH23, openai2024gpt4technicalreport, TheC3, geminiteam2024gemini15unlockingmultimodal} have gained  attention for combining visual perception with language generation. Refining VLMs with Reinforcement Learning \citep{DBLP:conf/nips/Ouyang0JAWMZASR22} improves alignment with human preferences. Recent success in Group Relative Policy Optimization (GRPO) \citep{deepseekai2025deepseekr1incentivizingreasoningcapability} has led to RL fine-tuned VLMs like AlphaDrive \citep{jiang2025alphadriveunleashingpowervlms}, VLM-R1 \citep{shen2025vlmr1stablegeneralizabler1style}, Praxis-VLM \cite{hu2025praxisvlmvisiongroundeddecisionmaking}, and MedVLM-R1 \citep{pan2025medvlmr1incentivizingmedicalreasoning}, broadening their applications. 

\textbf{Visually Impaired Assistance (VIA)} is a broad and diverse field
\cite{gao2025wearable, cai2024navigating, DBLP:conf/chi/GuanX024, DBLP:conf/chi/LiLKC24}. VIA with VLMs is closely related to visual captioning and Visual Question Answering (VQA). VIALM \citep{zhao2024vialmsurveybenchmarkvisually} frames VIA as a VQA task, generating guidance from environment images and user requests. While VIALM emphasizes environment-grounded guidance with tactile information, it is not specifically designed for navigation. WalkVLM \citep{yuan2025walkvlmaidvisuallyimpairedpeople} extends this to dynamic walking assistance and introduces the Walking Awareness Dataset (WAD). Though WalkVLM tackles navigation, its focus remains on video captioning rather than precise orientation and mobility guidance.

There are two main branches for NIG studies: NIG for embodied agents and NIG for the visually impaired. Prior research on \textbf{Navigation Instruction Generation (NIG) for embodied agents} has primarily focused on visual processing while generating trajectory-level instructions. BEVInstructor \citep{DBLP:conf/eccv/FanLWY24} employs a Bird's-Eye View encoder. SAS \citep{DBLP:conf/acl/GopinathanMAS24} uses semantic knowledge with adversarial reward learning. C-Instructor \citep{DBLP:conf/eccv/KongCWSHYL24} focuses on style-controlled instruction generation. Our proposed approach differs in two significant ways: (1) it prioritizes navigation feedback for VLM fine-tuning; and (2) it generates step-level in-situ instructions.

In the \textbf{NIG-VI} field, ASSISTER \citep{DBLP:conf/eccv/HuangSZBBO22} introduced the UrbanWalk benchmark and developed a navigation assistance model.  Our work offers two improvements: (1) we introduce a more detailed evaluation benchmark covering orientation, mobility, scene description, and safety warnings, informed by formative studies from the Human–Computer Interaction field that investigate the needs of VI users \citep{merchant2024generatingcontextuallyrelevantnavigationinstructions, zhao2025lessmorereducingcognitive, DBLP:conf/uist/ChangLG24}; and (2) we leverage advanced VLMs within a GRPO framework, leading to more effective instructions.
\section{Methodology}
Grounded in the \textbf{Speaker-Follower paradigm} \cite{DBLP:conf/nips/FriedHCRAMBSKD18}, the rationale for LaF-GRPO is to apply \textbf{Theory-of-Mind (ToM)} principles \cite{DBLP:conf/acl/ZhaoND23}. Our LLM-as-Follower simulates a user's cognitive mapping, generating feedback from their anticipated interpretation.
\begin{figure*}[t]
  \centering
  \includegraphics[width=\textwidth]{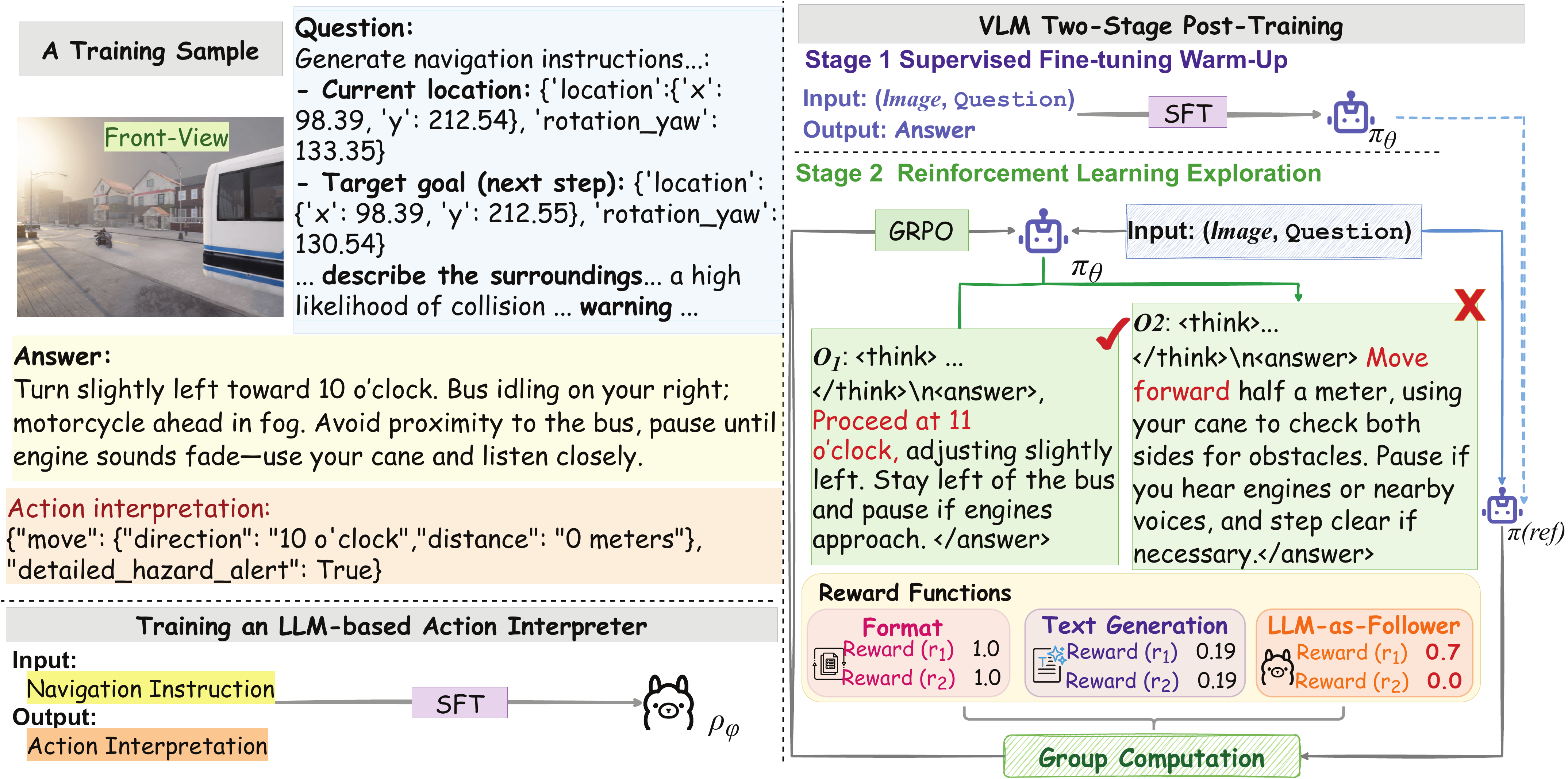}
  \caption{Method Overview. Top left: An training sample with input, target output, and action interpretation. Bottom left: Action interpreter training using \texttt{LLaMA-3-8B-Instruct} to simulate VI users' navigation responses. Right: Post-training for VLMs with LaF-GRPO. The LaF reward differentiates outputs when Format and Text Generation rewards are identical.}
  \label{fig:method}
\end{figure*}
\subsection{NIG-VI Task Formulation}
In the NIG-VI task, a VLM-based assistant generates in-situ, step-by-step natural language instructions to guide a visually impaired (VI) user along a pre-planned route $P = [p_1, \dots, p_K]$. The route $P$ consists of positional waypoints $p_i$ leading to a destination and is generated using the A* algorithm \cite{DBLP:conf/eccv/HuangSZBBO22}. At each discrete step $i$ of the navigation, the VLM receives two primary inputs: a front-view camera image $x_{\text{image}}^{(i)}$ and a task description which includes the user's current pose $x_{\text{pose}}^{(i)} = (x_{\text{loc}}^{(i)}, x_{\text{rot}}^{(i)})$ represented by their location $x_{\text{loc}}^{(i)} \in \mathbb{R}^3$ and rotation $x_{\text{rot}}^{(i)} \in \mathbb{R}^3$ within a global map coordinate system, as well as the next target waypoint $p_{i+1} \in P$. Based on these inputs, the VLM $\pi$ generates a sequence of tokens $y = y^{(i)} = (y_1^{(i)}, y_2^{(i)}, \dots, y_{t}^{(i)})$ of token length $t$. The generated instruction $y$ might also include details about the current surroundings captured in $x_{\text{image}}^{(i)}$ and any necessary safety alerts:
\begin{equation}
y_{j}\sim \pi_\theta(y_{j}^{(i)} | x_{\text{image}}^{(i)}, x_{\text{loc}}^{(i)}, x_{\text{rot}}^{(i)}, p_{i+1}, y_{<{j}}^{(i)})
\end{equation}
where $\theta$ denotes the adjustable model parameters.
\subsection{The LaF-GRPO Framework}
We propose LaF-GRPO to address the challenges of generating navigation instructions that are \textit{human-centered and practically usable for VI users}, while \textit{mitigating the need for costly real-world data collection with VI participants}. The overview of LaF-GRPO is illustrated in Figure \ref{fig:method}, where the framework comprises two key components: (1) an \textbf{action interpreter} LLM (without a visual encoder to \textit{`see}') that simulates VI users' responses to navigation instructions by interpreting how these users would act upon hearing the instructions, and (2) a VLM post-training procedure that generates these instructions with (1)'s feedback.

\textbf{Action Interpreter.} To simulate a VI user's response to navigation instructions, we use Supervised Fine-tuning (SFT) to train an LLM $\rho_{\varphi}$ to predict potential actions. Prior to its deployment, the interpreter is validated to have a parse accuracy above 98\% on a held-out set, ensuring its reliability for generating reward signals.  Given VLM-generated instruction tokens $y$, it produces a structured action interpretation $\mathcal{A}$. Formally, we define $\mathcal{A}$ as a structured dictionary containing: (1) a \textit{`move'} action with associated \textit{`direction'} (indicated using clock positions \cite{yuan2025walkvlmaidvisuallyimpairedpeople}) and \textit{`distance'} parameters, and (2) a \textit{`detailed\_hazard\_alert'} boolean flag that indicates whether the user perceives warnings about nearby obstacles, as illustrated in Figure \ref{fig:method} Left.

\textbf{Navigation Instruction Generator.} For VI guidance, we use a pre-trained VLM $\pi$ with parameters $\theta$ for in-situ navigation instruction generation. The training of this generator involves two stages: Supervised Fine-tuning (SFT) and Group Relative Policy Optimization (GRPO). Our proposed LaF-GRPO framework enhances standard GRPO (see below) by incorporating a novel LLM-as-Follower reward.

{\textit{GRPO.}}
The training process of GRPO aims to optimize the policy $\pi_{\theta}$ by maximizing the objective function $\mathcal{J}_{\text{GRPO}}(\theta)$. For a given query $q$, GRPO first samples a batch of $G$ outputs $\{o_1, o_2, \ldots, o_G\}$ using an older version of the policy, $\pi_{\theta_{\text{old}}}$. The training process of GRPO aims to optimize the policy $\pi_{\theta}$ by maximizing the objective function:
\begin{equation}
\mathcal{J}_{\text{GRPO}}(\theta) = \mathbb{E}_{q, \{o_i\} \sim \pi_{\theta_{\text{old}}}} \left[ \frac{1}{G} \sum_{i=1}^{G} \mathcal{L}_i - \beta \mathbb{D}_{\text{KL}}(\pi_{\theta} || \pi_{\text{ref}}) \right]
\end{equation}
Here, the term $\mathcal{L}_i$ represents the clipped surrogate objective used in PPO \citep{schulman2017proximalpolicyoptimizationalgorithms}:
\begin{equation}
    \mathcal{L}_i = \min(w_i A_i, \text{clip}(w_i, 1-\epsilon, 1+\epsilon)A_i)
\end{equation}
where $w_i = \frac{\pi_{\theta}(o_i|q)}{\pi_{\theta_{\text{old}}}(o_i|q)}$ is the importance sampling ratio, $A_i$ is the estimated advantage for the output $o_i$, based on relative rewards of the outputs inside each group only, calculated as $ A_i = \frac{r_i - \text{mean}(\{r_1, r_2, \ldots, r_G\})}{\text{std}(\{r_1, r_2, \ldots, r_G\})}$, and $\epsilon$ is a clipping hyperparameter. The second term, $-\beta \mathbb{D}_{\text{KL}}(\pi_{\theta} || \pi_{\text{ref}})$, regularizes the policy by penalizing divergence from a reference policy $\pi_{\text{ref}}$ with coefficient $\beta$. This regularization stabilizes training by keeping the model close to the original effective policy, preventing it from losing previously learned capabilities.
\subsection{LaF-GRPO Reward Functions}
LaF-GRPO utilizes three reward functions as follows.

\textbf{Format Reward.}
To encourage controllable generation, we adopt this binary reward ($r_{\text{format}} \in \{0,1\}$) that evaluates structural compliance with the expected response format. Here, the reward would equal 1 if the output follows the required format pattern \texttt{\seqsplit{`<think>.*?</think>\textbackslash n<answer>.*?</answer>'}} in sequence, and 0 otherwise.

\textbf{Text Generation Reward.}
The text generation reward ($r_{meteor}$) is calculated as the METEOR score between the output and the ground-truth reference. METEOR is selected based on its evaluation of semantic overlap, incorporating synonymy and stemming to provide a nuanced assessment.

\textbf{LLM-as-Follower Reward.} 
The LLM-as-Follower reward ($r_{LaF}$) assesses the navigational quality of generated instructions by comparing their interpreted actions (\textit{move direction}, \textit{distance}, and \textit{alert flag}) with those of a ground-truth reference. The rationale behind this design is that spatial factors, such as directional accuracy ($a_{dir}$) and movement distance precision ($a_{dist}$), play a direct and critical role in determining navigation success. In addition, safety alert flags ($a_{alert}$) serve as supplementary support for VI navigation by indicating potential hazards, though they are not primary determinants of success \citep{giudice2008blind, younis2019smart}. Accordingly, we formulate the reward as:
\begin{equation}
\begin{split}
r_{LaF} = & \, w_{dir} \, \delta(a_{dir}, a_{dir}^{ref}) + w_{dist} \, \delta(a_{dist}, a_{dist}^{ref}) \\
& + w_{alert} \, \delta(a_{alert}, a_{alert}^{ref})
\end{split}
\end{equation}
$\delta(\cdot)$ denotes an exact match. $r_{LaF}$ is in the range [0, 1].
\section{Benchmark: NIG4VI}
\begin{table*}[ht]
\small
\centering
\begin{tabular}{@{} p{5.2cm} l r c c c c c @{}}
\toprule
Benchmark & Level & \# Samples & VIA & NIG & Spatial Acc. & Open-ended & Open-sourced \\
\midrule
R2R \citep{DBLP:conf/cvpr/AndersonWTB0S0G18} & High & 21k & \ding{55} & \Checkmark & \ding{55} & \Checkmark & \Checkmark \\
REVERIE \citep{DBLP:conf/cvpr/QiW0WWSH20} & High & 10k / 6k & \ding{55} & \Checkmark & \ding{55} & \Checkmark & \Checkmark \\
\midrule
UrbanWalk \citep{DBLP:conf/eccv/HuangSZBBO22} & Detailed & 2.6k & \Checkmark & \Checkmark & \Checkmark & \ding{55} & \ding{55} \\
\citet{merchant2024generatingcontextuallyrelevantnavigationinstructions} & Detailed & 48 & \Checkmark & \Checkmark & \ding{55} & \Checkmark & \ding{55} \\
VIALM \citep{zhao2024vialmsurveybenchmarkvisually} & Detailed & 200 & \Checkmark & \ding{55} & \ding{55} & \Checkmark & \Checkmark \\
WAD \citep{yuan2025walkvlmaidvisuallyimpairedpeople} & Detailed & 12k / 120k & \Checkmark & \Checkmark & \ding{55} & \Checkmark & \Checkmark \\
\midrule
\textbf{NIG4VI (Ours)} & Detailed & 3k / 24k & \Checkmark & \Checkmark & \Checkmark & \Checkmark & \Checkmark \\
- w/o pre-calculation & Detailed & 1.5k / 12k & \Checkmark & \Checkmark & \Checkmark & \Checkmark & \Checkmark \\
- with pre-calculation & Detailed & 1.5k / 12k & \Checkmark & \Checkmark & \Checkmark & \Checkmark & \Checkmark \\
\bottomrule
\end{tabular}
\caption{Comparison of NIG4VI with existing benchmarks.}
\label{tab:benchmark_comparison}
\end{table*}
We introduce the NIG4VI benchmark to address the scarcity of benchmark resources in this field. Inspired by UrbanWalk, NIG4VI utilizes the open-sourced CARLA Simulator \citep{Dosovitskiy17} to collect samples from diverse scenarios, including varied environments and weather conditions. Pedestrian trajectories are generated using A* algorithm, with precise geospatial coordinates, orientation, frontal-view images, and semantic segmentation images recorded at each step. NIG4VI offers two advantages: (1) its use of a realistic coordinate system facilitates easier transfer to real-world GPS applications, and (2) it enables the cost-effective generation of accurate and extensive data. Table \ref{tab:benchmark_comparison} details NIG4VI's advantages over other datasets.

\begin{figure}
  \includegraphics[width=0.97\columnwidth]{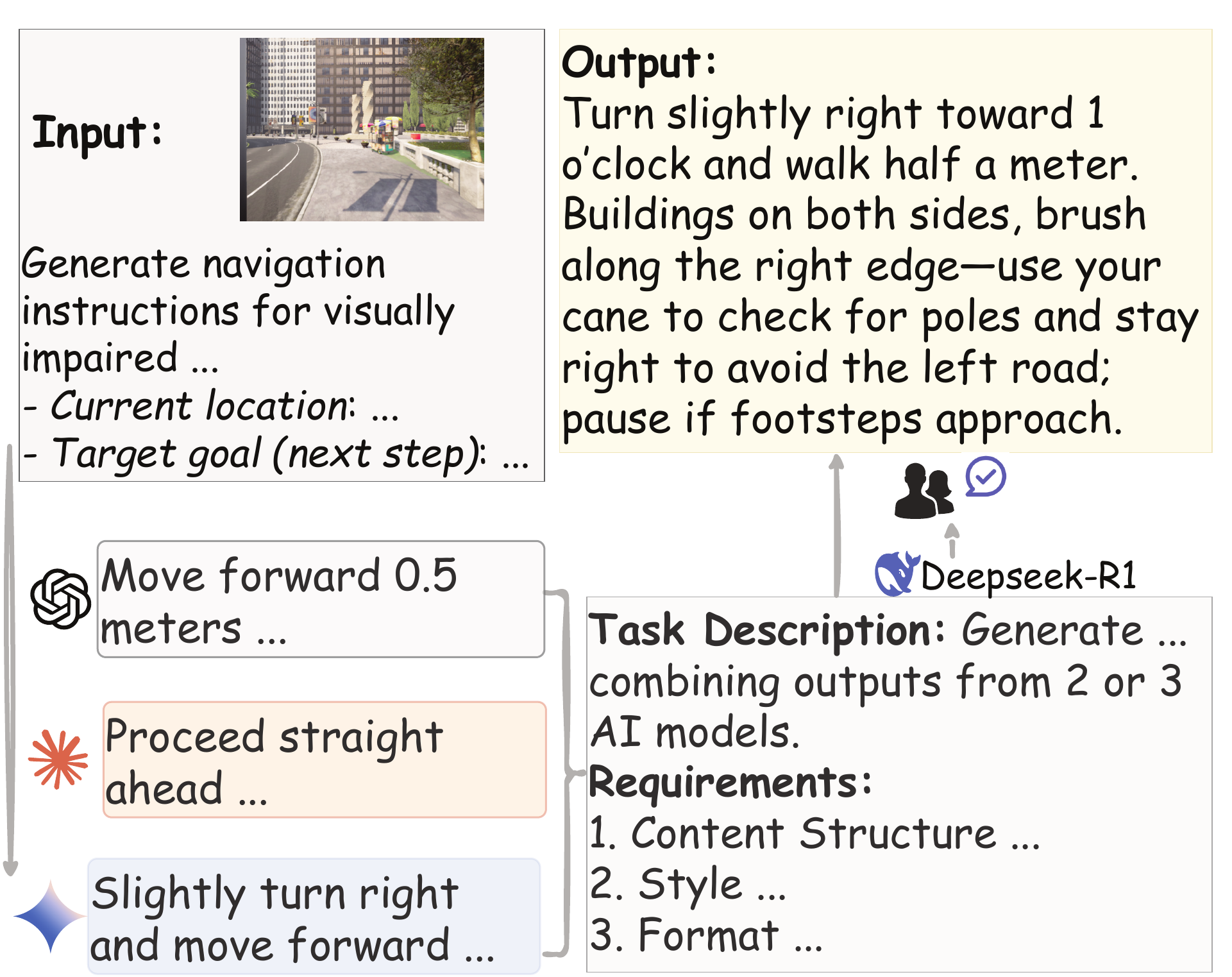}
  \caption{Dataset instances are first generated using Vision-R1's modality bridging method \citep{huang2025visionr1incentivizingreasoningcapability} and then reviewed and refined by human annotators.}
  \label{fig:nig_gt_gen}
\end{figure}

\textbf{Dataset Construction.} Each question's input includes the user's current location/rotation, the next step's location/orientation, and visual scene data. This information is then structured within a prompt template that includes a detailed task description. The synthesis of the output is a multi-stage process involving both advanced reasoning models and human annotation, as illustrated in Figure \ref{fig:nig_gt_gen}. Initially, several leading VLMs, specifically GPT-4o, Claude-3.5, and Gemini-2, generate predictions. Following a modality bridging approach, similar to that employed in Vision-R1 \citep{huang2025visionr1incentivizingreasoningcapability}, these outputs are processed through DeepSeek-R1 to enhance blindness-oriented spatial guidance and navigability. Crucially, all instructions undergo rigorous human verification. This task is carried out by two annotators, both proficient in English and holding at least an undergraduate-level education, following a similar practice in \citep{zhao2024vialmsurveybenchmarkvisually}. The verification involves a two-stage process: first, one annotator performs initial content adjustments, adhering to task requirements. Subsequently, the second annotator independently reviews and verifies this work. Throughout this process, the primary verification criteria include: (1) elimination of visual references (e.g., color-based descriptors), (2) validation of non-visual landmarks, and (3) confirmation of metric precision for mobility-critical parameters.

\textbf{Dataset Statistics.} On average, each town contributes approximately 26.2 navigation routes. The average Euclidean distance between the start and end points of these routes is 113.51 units, with an average of 353.8 steps. After deduplication, the dataset yielded an average of 2,222.7 step-level (image, question) samples per town. It is partitioned into a training set of 1,500 samples from Town01 and a test set. The test set comprises the remaining 613 \textbf{\textit{intra-town}} samples from Town01, along with all \textbf{\textit{inter-town}} samples from Town02 (2,579), Town03 (2,260), Town04 (2,316), Town05 (1,935), and Town10 (2,133). Each data sample is available in two versions: \textit{`with pre-calculation'} and \textit{`without pre-calculation'}. The \textit{`without pre-calculation'} version requires the VLM to independently calculate navigational parameters (e.g., distance, direction), presenting a greater challenge in guidance generation. Conversely, the \textit{`with pre-calculation'} version provides the VLM with basic mathematical movement information. The VLM must validate this data and assess the surroundings to generate the navigation instruction. 

\section{Experimental Settings}
\textbf{Dataset.} Experiments utilized the NIG4VI dataset, with Intra-town ($N=613$) and Inter-town ($N=11,223$) test subsets, under \textit{`with/without pre-calculation'} conditions.

\textbf{Models.} Diverse VLMs were evaluated, falling into two main groups: (1) remote models, such as \texttt{GPT-4o} \citep{openai2024gpt4ocard}, \texttt{\seqsplit{Claude-3-5-sonnet-20240620}} \citep{Anthropic2024bClaudeSonnet}, and \texttt{\seqsplit{Gemini-2.0-flash-thinking-exp-01-21}} \citep{GoogleDeepMind2024bGeminiIntroMisc}; and (2) smaller, locally runnable VLMs, including \texttt{DeepSeek-VL-7B} \citep{lu2024deepseekvl}, \texttt{LLaVA-v1.6-Mistral-7B} \citep{DBLP:conf/cvpr/LiuLLL24}, \texttt{MiniCPM-o-2.6-8B} \citep{yao2024minicpm}, \texttt{Intern-VL-2.5-8B} \citep{chen2024internvl}, and \texttt{Qwen2.5-VL-3B/7B} \citep{Qwen2.5-VL}.

\textbf{Baselines.}  We compare LaF-GRPO against two primary baseline methods:
(1)~\textbf{Zero-shot}: Models are applied directly to NIG4VI without prior fine-tuning.
(2)~\textbf{Supervised Fine-tuning (SFT)}: Models are fine-tuned to generate instructions. We implement two variants of LaF-GRPO:
(a)~\textbf{Zero-(LaF-GRPO)}: LaF-GRPO is applied directly to the base model without SFT.
(b)~\textbf{SFT+(LaF-GRPO)}: LaF-GRPO is applied to models that have first undergone SFT.

\textbf{Evaluation Metrics.} Following previous studies in NIG \citep{DBLP:conf/eccv/HuangSZBBO22, DBLP:conf/eccv/FanLWY24, DBLP:conf/eccv/KongCWSHYL24}, model performance was evaluated using a suite of widely adopted metrics: BLEU \citep{DBLP:conf/acl/PapineniRWZ02}, ROUGE \citep{lin-2004-rouge}, METEOR \citep{banerjee-lavie-2005-meteor}, and SPICE \citep{anderson2016spicesemanticpropositionalimage}. Human studies and LLM-as-Judge evaluations were conducted to assess navigational accuracy and user preference for instruction clarity and helpfulness.

\textbf{Implementation Details.} LaF-GRPO training utilized a single NVIDIA H20 GPU (96\,GB of memory). This hardware supports loading an 8B-param LLM (\texttt{LLaMA-3-8B}) and a 3B/7B-param Qwen2.5-VL model for LoRA \citep{hu2022lora} fine-tuning. The reward weights were configured as $(w_{dir}, w_{dist}, w_{alert}) = (0.4, 0.4, 0.2)$ based on analysis of navigation failure factors, prioritizing spatial parameters over contextual alerts. Training on $3k$ samples took about $15$ hours, with the hyperparameter group size $G$ set to $8$.
\section{Results and Discussions}
\begin{table*}[!htb]
\small
\setlength{\tabcolsep}{1pt}
\centering
\begin{tabular}{cclrrrrrrrr}
\toprule
\multirow{2}{*}{Pre-Cal.} & \multirow{2}{*}{Paradigm} & \multirow{2}{*}{Model} & \multicolumn{4}{c}{Intra-town ($N$ = 613)} & \multicolumn{4}{c}{Inter-town ($N$ =11,223)} \\
\cmidrule(lr){4-7} \cmidrule(lr){8-11}
& & & BLEU $\uparrow$ & ROUGE $\uparrow$ & METEOR $\uparrow$ & SPICE $\uparrow$ & BLEU $\uparrow$ & ROUGE $\uparrow$ & METEOR $\uparrow$ & SPICE $\uparrow$ \\
\midrule
\multirow{13}{*}{No} & \multirow{7}{*}{\textbf{Zero-Shot}} 
& DeepSeek-VL-7B & 2.179 & 0.152 & 0.182 & 0.116 & 2.223 & 0.157 & 0.196 & 0.112 \\
& & MiniCPM-o-8B & 2.009 & 0.145 & 0.234 & 0.131 & 1.969 & 0.142 & 0.233 & 0.129 \\
& & Intern-VL-8B & 1.448 & 0.150 & 0.215 & 0.126 & 1.517 & 0.149 & 0.216 & 0.120 \\
& & LLaVA-7B & 1.021 & 0.103 & 0.201 & 0.111 & 1.037 & 0.107 & 0.206 & 0.109 \\
& & \fbox{Qwen-VL-7B} & 3.204 & 0.202 & 0.211 & 0.166 & 3.128 & 0.194 & 0.210 & 0.157 \\
& & GPT-4o & 1.748 & 0.169 & 0.249 & 0.149 & 1.617 & 0.165 & 0.249 & 0.142 \\
& & Claude-3.5 & 2.803 & 0.216 & 0.304 & 0.211 & 2.749 & 0.211 & 0.301 & 0.202 \\
& & Gemin-2 & 4.105 & 0.236 & 0.232 & 0.232 & 4.422 & 0.252 & 0.238 & 0.236 \\
\cmidrule(lr){2-11}
& \multirow{2}{*}{\makecell{\textbf{Zero-}\\\textbf{(LaF-GRPO)}}}& \fbox{Qwen-VL-3B} & \fbox{3.292} & 0.230 & 0.248 & \fbox{0.230} & \fbox{3.972} & \fbox{0.255} & 0.259 & \fbox{0.244} \\
& & \fbox{Qwen-VL-7B} & 3.272 & \fbox{0.234} & \fbox{0.256} & 0.222 & 3.566 & 0.252 & \fbox{0.260} & 0.227 \\
\cmidrule(lr){2-11}
& \multirow{2}{*}{\textbf{SFT}} & Qwen-VL-3B & 9.099 & 0.282 & 0.496 & 0.274 & 8.949 & 0.284 & 0.500 & 0.276 \\
& & Qwen-VL-7B & 9.937 & \underline{0.291} & 0.518 & \underline{0.275} & \underline{9.709} & \underline{0.294} & 0.526 & \textbf{0.281} \\
\cmidrule(lr){2-11}
& \multirow{2}{*}{\makecell{\textbf{SFT+}\\\textbf{(LaF-GRPO)}}} & Qwen-VL-3B & \textbf{10.921} & \textbf{0.323} & \underline{0.528} & 0.274 & \textbf{10.157} & \textbf{0.309} & \underline{0.527} & 0.276 \\
& & Qwen-VL-7B & \underline{10.037} & 0.284 & \textbf{0.545} & \textbf{0.283} & 9.002 & 0.276 & \textbf{0.535} & \underline{0.278} \\
\midrule
\multirow{13}{*}{Yes} & \multirow{7}{*}{\textbf{Zero-Shot}} & DeepSeek-VL-7B & 2.517 & 0.170 & 0.224 & 0.161 & 2.600 & 0.173 & 0.237 & 0.161 \\
& & MiniCPM-o-8B & 2.349 & 0.166 & 0.210 & 0.136 & 2.517 & 0.177 & 0.220 & 0.144 \\
& & Intern-VL-8B & 1.496 & 0.132 & 0.233 & 0.133 & 1.517 & 0.134 & 0.238 & 0.132 \\
& & LLaVA-7B & 1.284 & 0.120 & 0.222 & 0.131 & 1.285 & 0.121 & 0.229 & 0.127 \\
& & \fbox{Qwen-VL-7B} & 2.903 & 0.188 & 0.231 & 0.178 & 3.080 & 0.194 & 0.243 & 0.180 \\
& & GPT-4o & 2.766 & 0.204 & 0.302 & 0.198 & 2.967 & 0.213 & 0.323 & 0.211 \\
& & Claude-3.5 & 4.124 & 0.236 & 0.349 & 0.257 & 3.400 & 0.214 & 0.326 & 0.224 \\
& & Gemin-2 & 5.132 & 0.252 & 0.266 & 0.269 & 6.144 & 0.276 & 0.283 & 0.284 \\
\cmidrule(lr){2-11}
& \multirow{2}{*}{\makecell{\textbf{Zero-}\\\textbf{(LaF-GRPO)}}} & \fbox{Qwen-VL-3B} & \fbox{3.798} & \fbox{0.249} & 0.280 & \fbox{0.261} & \fbox{4.584} & \fbox{0.271} & \fbox{0.288} & \fbox{0.274} \\
& & \fbox{Qwen-VL-7B} & 3.678 & 0.241 & \fbox{0.281} & 0.229 & 4.284 & 0.262 & 0.286 & 0.230 \\
\cmidrule(lr){2-11}
& \multirow{2}{*}{\textbf{SFT}} & Qwen-VL-3B & 9.923 & \underline{0.308} & 0.512 & 0.280 & \underline{10.724} & \underline{0.318} & 0.519 & 0.280 \\
& & Qwen-VL-7B & 9.639 & 0.270 & 0.521 & 0.283 & 9.710 & 0.272 & 0.524 & \underline{0.287} \\
\cmidrule(lr){2-11}
& \multirow{2}{*}{\makecell{\textbf{SFT+}\\\textbf{(LaF-GRPO)}}} & Qwen-VL-3B & \textbf{11.727} & \textbf{0.342} & \underline{0.541} & \underline{0.286} & \textbf{10.813} & \textbf{0.333} & \underline{0.535} & 0.279 \\
& & Qwen-VL-7B & \underline{10.499} & 0.285 & \textbf{0.556} & \textbf{0.292} & 9.232 & 0.275 & \textbf{0.542} & \textbf{0.288} \\
\bottomrule
\end{tabular}
\caption{Evaluation results on NIG4VI across Intra-town and Inter-town subsets. \textbf{Bold} values represent the highest score for each metric under a specific setting (with / without pre-calculation), while \underline{underlined} values indicate the second-highest score. \fbox{Boxed} values highlight the best performing Qwen2.5-VL model within the Zero-Shot and Zero-(LaF-GRPO) categories. }
\label{tab:nig4vi_results}
\end{table*}

\begin{figure*}[ht]
  \centering
  \includegraphics[width=\textwidth]{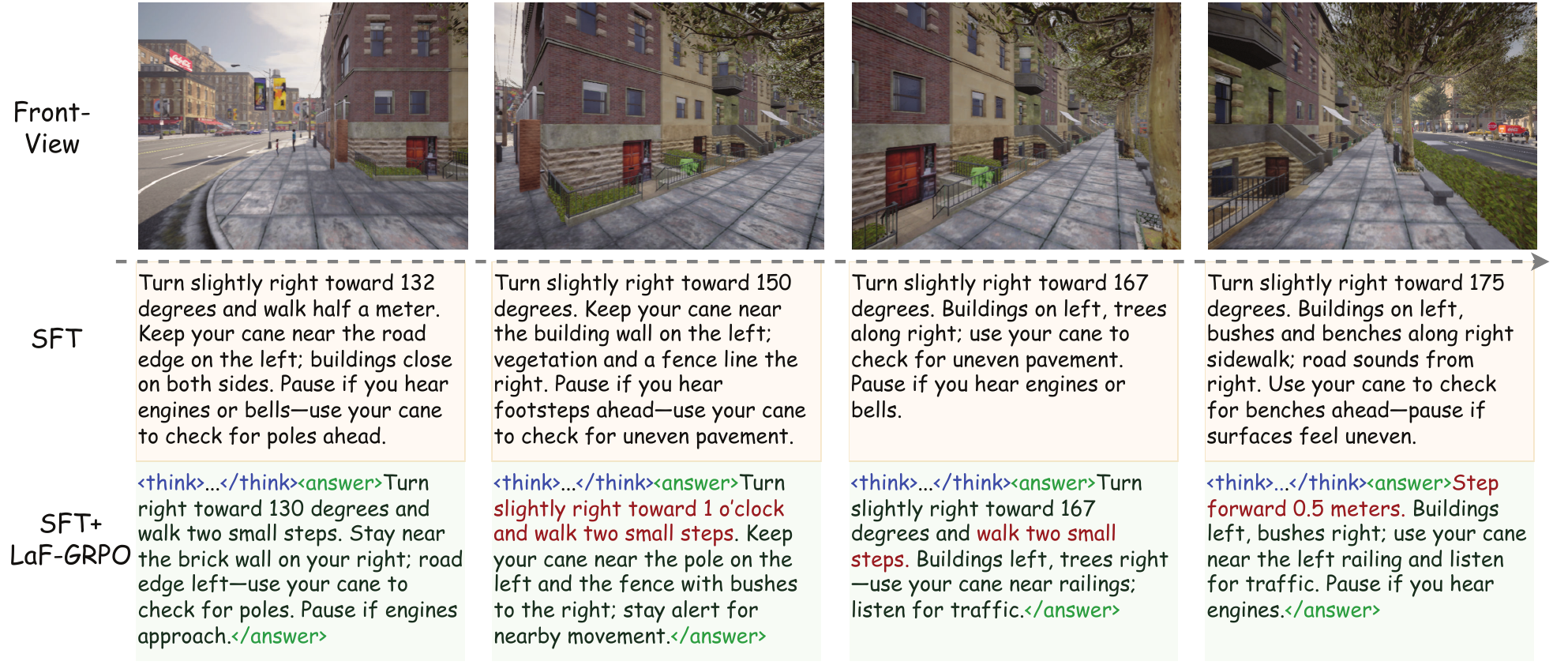}
  \caption{A comparative case study of navigational guidance provided by SFT and SFT+(LaF-GRPO) methods.}
  \label{fig:case_study}
\end{figure*}
\begin{table*}[!htb]
\centering
\small
\setlength{\tabcolsep}{2.5pt}
\begin{tabular}{lcccccccccccc}
\toprule
\multirow{2}{*}{\begin{tabular}[c]{@{}l@{}}Pre-Cal.\end{tabular}} & \multicolumn{3}{c}{Reward Types} & \multicolumn{4}{c}{Intra-town ($N$ = 613)} & \multicolumn{4}{c}{Inter-town ($N$ = 11,223)} \\
\cmidrule(lr){2-4} \cmidrule(lr){5-8} \cmidrule(lr){9-12}
 & \textbf{Format} & \textbf{Meteor} & \textbf{LLM} & BLEU $\uparrow$ & ROUGE $\uparrow$ & METEOR $\uparrow$ & SPICE $\uparrow$ & BLEU $\uparrow$ & ROUGE  $\uparrow$ & METEOR $\uparrow$ & SPICE $\uparrow$ \\
\midrule
\multirow{3}{*}{No} 
 & $\checkmark$ & & & 10.251 & 0.318 & 0.524 & 0.278 & 9.401 & 0.304 & 0.523 & 0.279 \\
 & $\checkmark$ & $\checkmark$ & & 10.912 & 0.317 & 0.525 & \textbf{0.279} & 10.076 & 0.306 & 0.521 & \textbf{0.279} \\
 & $\checkmark$ & $\checkmark$ & $\checkmark$ & \textbf{10.921} & \textbf{0.323} & \textbf{0.528} & 0.274 & \textbf{10.157} & \textbf{0.309} & \textbf{0.527} & 0.276 \\
\midrule
\multirow{3}{*}{Yes} 
 & $\checkmark$ & & & 11.269 & 0.337 & 0.538 & \textbf{0.292} & 10.217 & 0.328 & 0.530 & \textbf{0.282} \\
 & $\checkmark$ & $\checkmark$ & & 11.602 & 0.339 & 0.539 & 0.284 & 10.753 & 0.331 & 0.531 & 0.280 \\
 & $\checkmark$ & $\checkmark$ & $\checkmark$ & \textbf{11.727} & \textbf{0.342} & \textbf{0.541} & 0.286 & \textbf{10.813} & \textbf{0.333} & \textbf{0.535} & 0.280 \\
\bottomrule
\end{tabular}
\caption{Ablation study results for the Qwen2.5-VL-3B model on the NIG4VI dataset with different \textbf{reward functions}.}

\label{ablash_reward_types}
\end{table*}

\begin{table*}[!htb]
\small
\setlength{\tabcolsep}{2pt}
\centering
\begin{tabular}{@{}c>{\centering\arraybackslash}ccccccccc@{}}
\toprule
\multirow{2}{*}{\begin{tabular}[c]{@{}c@{}}Pre-Cal.\end{tabular}} & \multirow{2}{*}{Model} & \multicolumn{4}{c}{Intra-town ($N$ = 613)} & \multicolumn{4}{c}{Inter-town ($N$ = 11,223)} \\
\cmidrule(lr){3-6} \cmidrule(l){7-10}
 & & BLEU $\uparrow$ & ROUGE $\uparrow$ & METEOR $\uparrow$ & SPICE $\uparrow$ & BLEU $\uparrow$ & Rouge $\uparrow$ & METEOR $\uparrow$ & SPICE $\uparrow$ \\
\midrule
\multirow{3}{*}{No} 
& 7B-format+meteor+LLM (\textbf{1k}) & 9.401 & 0.283 & 0.529 & 0.274 & 8.963 & 0.281 & 0.530 & 0.275 \\
& 7B-format+meteor+LLM (\textbf{2k}) & 9.657 & 0.280 & 0.539 & 0.276 & 9.001 & \textbf{0.276} & \textbf{0.535} & 0.274 \\
& 7B-format+meteor+LLM (\textbf{3k}) & \textbf{10.037} & \textbf{0.284} & \textbf{0.545} & \textbf{0.283} & \textbf{9.002} & \textbf{0.276} & \textbf{0.535} & \textbf{0.278} \\
\midrule
\multirow{3}{*}{Yes} 
& 7B-format+meteor+LLM (\textbf{1k}) & 10.265 & 0.279 & 0.543 & 0.286 & 9.463 & 0.271 & 0.540 & 0.285 \\
& 7B-format+meteor+LLM (\textbf{2k}) & 10.136 & 0.284 & 0.550 & \textbf{0.292} & \textbf{9.245} & \textbf{0.276} & 0.541 & 0.284 \\
& 7B-format+meteor+LLM (\textbf{3k}) & \textbf{10.499} & \textbf{0.285} & \textbf{0.556} & \textbf{0.292} & 9.232 & 0.275 & \textbf{0.542} & \textbf{0.288} \\
\bottomrule
\end{tabular}
\caption{Ablation study results for the Qwen2.5-VL-7B model on the NIG4VI dataset with varying training \textbf{sample sizes}.}
\label{ablash_sample_num}
\end{table*}

\subsection{Main Results}
Table~\ref{tab:nig4vi_results} summarizes model performance on NIG4VI, categorized by pre-calculation and training paradigms, and evaluated on intra-town and inter-town subsets. Comparing LaF-GRPO with the baselines reveals: (1) \textbf{Zero-Shot vs. Zero-(LaF-GRPO):} Zero-(LaF-GRPO) significantly enhances the Zero-Shot performance of VLMs, validating the effectiveness of LaF-GRPO. While the Zero-(LaF-GRPO) results suggest that increased model size (from 3B to 7B) does not necessarily guarantee improved performance across all metrics, it is noteworthy that for METEOR evaluations, specifically in intra-town scenarios, the 7B model achieved the highest scores (i.e., $0.256$ and $0.281$). This outcome may be attributable to the use of METEOR as a text generation reward during training and to the potentially more refined tuning applied to the 7B models. (2) \textbf{SFT \& SFT+(LaF-GRPO):} SFT and SFT+(LaF-GRPO) yield significantly superior performance compared to Zero-Shot and Zero-(LaF-GRPO) models across all metrics and subsets, affirming the efficacy of fine-tuning. The SFT+(LaF-GRPO) approach further enhances performance beyond SFT. Moreover, under the SFT+(LaF-GRPO) paradigm, Qwen-VL-3B consistently achieves the highest BLEU and ROUGE scores, while Qwen-VL-7B excels in METEOR and SPICE. This performance pattern is observed for both intra-town and inter-town subsets and holds true regardless of pre-calculation. This may be attributable to 7B models demonstrating enhanced linguistic diversity in their outputs relative to 3B models. (3) \textbf{Additional Observations:} Scores improved with pre-calculation, which reduced computational difficulty, whereas the more challenging `w/o pre-calculation' setting caused more failures. Intra-town scores were higher than inter-town, likely due to a closer data distribution with the training set. We also observed that LaF-GRPO generated significantly more concise, user-friendly instructions (e.g., LaF-GRPO-7B: ~34.1 tokens vs. GPT-4o: ~117.9 tokens).

\subsection{Ablation Studies}
\textbf{Reward Types.} Table~\ref{ablash_reward_types} presents an ablation study investigating the impact of different reward types during SFT+(LaF-GRPO) training with the Qwen-VL-3B model. LaF-GRPO, incorporating the LLM-as-Follower reward, consistently achieves the highest BLEU, ROUGE, and METEOR scores. This trend holds true across both intra-town and inter-town evaluations, with or without pre-calculation. This underscores the significant benefit of the LaF reward.

\textbf{Training Sample Sizes.} Table~\ref{ablash_sample_num} presents an ablation study on the Qwen2.5-VL-7B model trained with SFT+(LaF-GRPO), illustrating the effect of varying training sample sizes ($1k$, $2k$, and $3k$). For comprehensive metrics such as METEOR and SPICE, performance generally scales with the volume of training data. Across the majority of evaluated conditions, scaling up to $3k$ samples typically yields the optimal or near-optimal scores. Nevertheless, training with $2k$ samples also achieves comparable METEOR scores, indicating training data efficiency at this sample size.

\textbf{LaF-GRPO vs. Standard GRPO.} We conducted \textbf{LLM-as-Judge} and \textbf{human evaluations} on the inter-town subset. The LLM-as-Judge evaluation with GPT-4o showed LaF-GRPO's superior navigational accuracy (68.1\% vs. 67.3\%) and a greater preference rate for instruction clarity and helpfulness (58.3\% vs. 41.7\%). These findings were corroborated by a human preference study on 100 instruction pairs, evaluated by two trained graduate students acting as VI user simulators. To isolate instruction quality, the evaluators were instructed to navigate relying solely on text and ignoring visual cues. Their assessments achieved substantial inter-rater agreement (Cohen’s $\kappa $= 0.83), and the results confirmed a strong preference for LaF-GRPO instructions (76\% vs. 24\%) and its higher navigational accuracy (79.0\% vs. 77.5\%). We identify two main advantages of LaF-GRPO over standard GRPO, which utilizes only format and text generation rewards: (1)~\textbf{Navigational Accuracy:} LaF-GRPO provides more precise movement and orientation guidance. (2)~\textbf{Instruction Clarity:} The action interpreter component encourages VLMs to generate instructions that are clearer, more structured, and more comprehensible.

\subsection{Case Study}
Figure~\ref{fig:case_study} provides a qualitative comparison of SFT+(LaF-GRPO) against the SFT baseline. Notably, SFT+(LaF-GRPO) generates instructions with greater linguistic variety and more intuitive directional cues. For instance, in Step 2, SFT+(LaF-GRPO) employs an o'clock direction (\textit{``Turn slightly right toward 1 o'clock"}) and a relatable distance (\textit{"two small steps"}), contrasting with SFT's numerical bearing (\textit{"150 degrees"}). It can yield guidance that is more naturally understood by VI users. Furthermore, SFT+(LaF-GRPO), leveraging its internal reasoning process (i.e. the \texttt{<think>...</think>} blocks), frequently incorporates more environmental details and safety considerations. For example, its instruction for Step 4 (\textit{"Step forward 0.5 meters; ...use your cane near the left ... and listen for traffic"}) also emphasizes immediate safety interactions.

\subsection{Limitations and Future Work}
This study's reliance on simulation and proxy users, while necessary for large-scale, controlled, and safe initial testing, introduces limitations. Future work could explore real-world dataset collection and direct engagement with VI users.
\section{Conclusion}
This study addresses navigation instruction generation for the visually impaired individuals (NIG-VI). We constructed the NIG4VI benchmark. Following this, we developed LaF-GRPO, a novel training paradigm for VLMs that incorporates an LLM-as-Follower reward. Experimental evaluations established LaF-GRPO's superiority over baselines and standard GRPO, with qualitative analysis confirming the practicality of the generated instructions in real-world scenarios. We hope our work and the benchmark inspire the development of more effective aids for the visually impaired.
\section*{Acknowledgments}
This work was supported by the Innovation and Technology Fund (Project No. PRP/047/22FX), a grant from the Research Grants Council of the Hong Kong Special Administrative Region, China (Project No. PolyU/25200821), and PolyU Internal Fund from RC-DSAI (Project No. 1-CE1E). We also thank the reviewers for their constructive comments.
\bibliography{aaai2026}

@misc{hu2025praxisvlmvisiongroundeddecisionmaking,
      title={Praxis-VLM: Vision-Grounded Decision Making via Text-Driven Reinforcement Learning}, 
      author={Zhe Hu and Jing Li and Zhongzhu Pu and Hou Pong Chan and Yu Yin},
      year={2025},
      eprint={2503.16965},
      archivePrefix={arXiv},
      primaryClass={cs.CL},
      url={https://arxiv.org/abs/2503.16965}, 
}

@misc{zhao2025lessmorereducingcognitive,
      title={"Less is More": Reducing Cognitive Load and Task Drift in Real-Time Multimodal Assistive Agents for the Visually Impaired}, 
      author={Yi Zhao and Siqi Wang and Qiqun Geng and Erxin Yu and Jing Li},
      year={2025},
      eprint={2511.00945},
      archivePrefix={arXiv},
      primaryClass={cs.HC},
      url={https://arxiv.org/abs/2511.00945}, 
}

@inproceedings{DBLP:conf/uist/ChangLG24,
  author       = {Ruei{-}Che Chang and
                  Yuxuan Liu and
                  Anhong Guo},
  editor       = {Lining Yao and
                  Mayank Goel and
                  Alexandra Ion and
                  Pedro Lopes},
  title        = {WorldScribe: Towards Context-Aware Live Visual Descriptions},
  booktitle    = {Proceedings of the 37th Annual {ACM} Symposium on User Interface Software
                  and Technology, {UIST} 2024, Pittsburgh, PA, USA, October 13-16, 2024},
  pages        = {140:1--140:18},
  publisher    = {{ACM}},
  year         = {2024},
  url          = {https://doi.org/10.1145/3654777.3676375},
  doi          = {10.1145/3654777.3676375},
  bibsource    = {dblp computer science bibliography, https://dblp.org}
}

@inproceedings{DBLP:conf/nips/FriedHCRAMBSKD18,
  author       = {Daniel Fried and
                  Ronghang Hu and
                  Volkan Cirik and
                  Anna Rohrbach and
                  Jacob Andreas and et al.},
  editor       = {Samy Bengio and
                  Hanna M. Wallach and
                  Hugo Larochelle and
                  Kristen Grauman and
                  Nicol{\`{o}} Cesa{-}Bianchi and
                  Roman Garnett},
  title        = {Speaker-Follower Models for Vision-and-Language Navigation},
  booktitle    = {Advances in Neural Information Processing Systems 31: Annual Conference
                  on Neural Information Processing Systems 2018, NeurIPS 2018},
  pages        = {3318--3329},
  year         = {2018},
  url          = {https://proceedings.neurips.cc/paper/2018/hash/6a81681a7af700c6385d36577ebec359-Abstract.html},
  bibsource    = {dblp computer science bibliography, https://dblp.org}
}

@inproceedings{DBLP:conf/cvpr/LiuLLL24,
  author       = {Haotian Liu and
                  Chunyuan Li and
                  Yuheng Li and
                  Yong Jae Lee},
  title        = {Improved Baselines with Visual Instruction Tuning},
  booktitle    = {{IEEE/CVF} Conference on Computer Vision and Pattern Recognition,
                  {CVPR} 2024, Seattle, WA, USA},
  pages        = {26286--26296},
  publisher    = {{IEEE}},
  year         = {2024}
}

@inproceedings{DBLP:conf/chi/LiLKC24,
  author       = {Franklin Mingzhe Li and
                  Michael Xieyang Liu and
                  Shaun K. Kane and
                  Patrick Carrington},
  title        = {A Contextual Inquiry of People with Vision Impairments in Cooking},
  booktitle    = {Proceedings of the {CHI} Conference on Human Factors in Computing Systems, {CHI} 2024, Honolulu, HI, USA},
  pages        = {38:1--38:14},
  publisher    = {{ACM}},
  year         = {2024}
}

@inproceedings{DBLP:conf/chi/GuanX024,
  author       = {Zhitong Guan and
                  Zeyu Xiong and
                  Mingming Fan},
  title        = {FetchAid: Making Parcel Lockers More Accessible to Blind and Low Vision
                  People With Deep-learning Enhanced Touchscreen Guidance, Error-Recovery
                  Mechanism, and AR-based Search Support},
  booktitle    = {Proceedings of the {CHI} Conference on Human Factors in Computing
                  Systems, {CHI} 2024, Honolulu, HI, USA},
  pages        = {39:1--39:15},
  publisher    = {{ACM}},
  year         = {2024}
}

@misc{WHO2019vision,
  author       = {{World Health Organization}},
  title        = {World report on vision},
  year         = {2019},
  howpublished = {\url{https://www.who.int/publications/i/item/9789241516570}},
  note         = {Accessed: 2025-04-28}
}

@article{gao2025wearable,
  title={A wearable obstacle avoidance device for visually impaired individuals with cross-modal learning},
  author={Gao, Yun and Wu, Dan and Song, Jie and Zhang, Xueyi and et al.},
  journal={Nature Communications},
  volume={16},
  number={1},
  pages={2857},
  year={2025}
}

@inproceedings{DBLP:conf/acl/ZhaoND23,
  author       = {Lingjun Zhao and
                  Khanh Nguyen and
                  Hal Daum{\'{e}} III},
  title        = {Define, Evaluate, and Improve Task-Oriented Cognitive Capabilities
                  for Instruction Generation Models},
  booktitle    = {Findings of the Association for Computational Linguistics: {ACL} 2023,
                  Toronto, Canada},
  pages        = {3688--3706},
  publisher    = {Association for Computational Linguistics},
  year         = {2023}
}

@inproceedings{DBLP:conf/naacl/DouP22,
  author       = {Zi{-}Yi Dou and
                  Nanyun Peng},
  title        = {{FOAM:} {A} Follower-aware Speaker Model For Vision-and-Language Navigation},
  booktitle    = {Proceedings of the 2022 Conference of the North American Chapter of
                  the Association for Computational Linguistics: Human Language Technologies,
                  {NAACL} 2022, Seattle, WA, United States},
  pages        = {4332--4340},
  publisher    = {Association for Computational Linguistics},
  year         = {2022}
}

@inproceedings{cai2024navigating,
  author    = {Cai, Shaojun and Ram, Ashwin and Gou, Zhengtai and et al.},
  title     = {Navigating real-world challenges: A quadruped robot guiding system for visually impaired people in diverse environments},
  booktitle = {Proceedings of the 2024 CHI Conference on Human Factors in Computing Systems},
  pages     = {1--18},
  year      = {2024},
  publisher = {ACM},
  address   = {Honolulu, HI, USA}
}

@inproceedings{Dosovitskiy17,
  title = {{CARLA}: {An} Open Urban Driving Simulator},
  author = {Alexey Dosovitskiy and German Ros and Felipe Codevilla and Antonio Lopez and Vladlen Koltun},
  booktitle = {Proceedings of the 1st Annual Conference on Robot Learning},
  pages = {1--16},
  year = {2017}
}

@inproceedings{devlin-etal-2019-bert,
    title = "{BERT}: Pre-training of Deep Bidirectional Transformers for Language Understanding",
    author = "Devlin, Jacob  and
      Chang, Ming-Wei  and
      Lee, Kenton  and
      Toutanova, Kristina",
    booktitle = "Proceedings of the 2019 Conference of the North {A}merican Chapter of the Association for Computational Linguistics: Human Language Technologies, Volume 1 (Long and Short Papers)",
    month = jun,
    year = "2019",
    address = "Minneapolis, Minnesota",
    publisher = "Association for Computational Linguistics"
}

@inproceedings{hu2022lora,
  author       = {Edward J. Hu and
                  Yelong Shen and
                  Phillip Wallis and
                  Zeyuan Allen{-}Zhu and
                  Yuanzhi Li and
                  Shean Wang and
                  Lu Wang and
                  Weizhu Chen},
  title        = {LoRA: Low-Rank Adaptation of Large Language Models},
  booktitle    = {{ICLR} 2022},
  publisher    = {OpenReview.net},
  year         = {2022}
}

@misc{anderson2016spicesemanticpropositionalimage,
      title={SPICE: Semantic Propositional Image Caption Evaluation}, 
      author={Peter Anderson and Basura Fernando and Mark Johnson and Stephen Gould},
      year={2016},
      eprint={1607.08822},
      archivePrefix={arXiv},
}

@inproceedings{lin-2004-rouge,
    title = "{ROUGE}: A Package for Automatic Evaluation of Summaries",
    author = "Lin, Chin-Yew",
    booktitle = "Text Summarization Branches Out",
    month = jul,
    year = "2004",
    address = "Barcelona, Spain",
    publisher = "Association for Computational Linguistics"
}

@inproceedings{banerjee-lavie-2005-meteor,
    title = "{METEOR}: An Automatic Metric for {MT} Evaluation with Improved Correlation with Human Judgments",
    author = "Banerjee, Satanjeev  and
      Lavie, Alon",
    booktitle = "Proceedings of the {ACL} Workshop on Intrinsic and Extrinsic Evaluation Measures for Machine Translation and/or Summarization",
    month = jun,
    year = "2005",
    address = "Ann Arbor, Michigan",
    publisher = "Association for Computational Linguistics"
}

@inproceedings{DBLP:conf/acl/PapineniRWZ02,
  author       = {Kishore Papineni and
                  Salim Roukos and
                  Todd Ward and
                  Wei{-}Jing Zhu},
  title        = {Bleu: a Method for Automatic Evaluation of Machine Translation},
  booktitle    = {Proceedings of the 40th Annual Meeting of the Association for Computational
                  Linguistics, July 6-12, 2002, Philadelphia, PA, {USA}},
  pages        = {311--318},
  publisher    = {{ACL}},
  year         = {2002}
}

@misc{Anthropic2024bClaudeSonnet,
  author       = {{Anthropic}},
  title        = {Claude 3.5 Sonnet},
  year         = {2024},
  howpublished = {\url{https://www.anthropic.com/news/claude-3-5-sonnet}},
  note         = {Accessed: 2025-04-26}
}

@misc{GoogleDeepMind2024bGeminiIntroMisc,
  author       = {{Google DeepMind}},
  title        = {Introducing Gemini 2.0: Our New AI Model for the Agentic Era},
  year         = {2024},
  month        = {dec},
  howpublished = {\url{https://blog.google/technology/google-deepmind/google-gemini-ai-update-december-2024}},
  note         = {Accessed: 2025-04-21}
}

@misc{openai2024gpt4ocard,
      title={GPT-4o System Card}, 
      author={OpenAI},
      year={2024},
      eprint={2410.21276},
      archivePrefix={arXiv}
}

@misc{lu2024deepseekvl,
      title={DeepSeek-VL: Towards Real-World Vision-Language Understanding}, 
      author={Haoyu Lu and Wen Liu and Bo Zhang and et al.},
      year={2024},
      eprint={2403.05525},
      archivePrefix={arXiv},
      primaryClass={cs.AI}
}

@misc{yao2024minicpm,
  title={MiniCPM-V: A GPT-4V Level MLLM on Your Phone},
  author={Yao, Yuan and Yu, Tianyu and Zhang, Ao and Wang, Chongyi and Cui, Junbo and Zhu, Hongji and Cai, Tianchi and Li, Haoyu and Zhao, Weilin and He, Zhihui and others},
    year= {2024},
    eprint= {2408.01800},
archivePrefix = {arXiv}
}

@misc{Qwen2.5-VL,
  title={Qwen2.5-VL Technical Report},
  author={Bai, Shuai and Chen, Keqin and Liu, Xuejing and Wang, Jialin and Ge, Wenbin and Song, Sibo and et al.},
    year= {2025},
    eprint= {2502.13923},
    archivePrefix = {arXiv}
}

@inproceedings{chen2024internvl,
  title={Internvl: Scaling up vision foundation models and aligning for generic visual-linguistic tasks},
  author={Chen, Zhe and Wu, Jiannan and Wang, Wenhai and Su, Weijie and Chen, Guo and others},
  booktitle={Proceedings of the IEEE/CVF Conference on Computer Vision and Pattern Recognition},
  pages={24185--24198},
  year={2024}
}

@misc{huang2025visionr1incentivizingreasoningcapability,
      title={Vision-R1: Incentivizing Reasoning Capability in Multimodal Large Language Models}, 
      author={Wenxuan Huang and Bohan Jia and Zijie Zhai and Shaosheng Cao and Zheyu Ye and Fei Zhao and Zhe Xu and Yao Hu and Shaohui Lin},
      year={2025},
      eprint={2503.06749},
      archivePrefix={arXiv}
}

@misc{schulman2017proximalpolicyoptimizationalgorithms,
      title={Proximal Policy Optimization Algorithms}, 
      author={John Schulman and Filip Wolski and Prafulla Dhariwal and Alec Radford and Oleg Klimov},
      year={2017},
      eprint={1707.06347},
      archivePrefix={arXiv},
}

@misc{shen2025vlmr1stablegeneralizabler1style,
      title={VLM-R1: A Stable and Generalizable R1-style Large Vision-Language Model}, 
      author={Haozhan Shen and Peng Liu and Jingcheng Li and Chunxin Fang and Yibo Ma and Jiajia Liao and Qiaoli Shen and Zilun Zhang and Kangjia Zhao and Qianqian Zhang and Ruochen Xu and Tiancheng Zhao},
      year={2025},
      eprint={2504.07615},
      archivePrefix={arXiv}
}

@misc{pan2025medvlmr1incentivizingmedicalreasoning,
      title={MedVLM-R1: Incentivizing Medical Reasoning Capability of Vision-Language Models (VLMs) via Reinforcement Learning}, 
      author={Jiazhen Pan and Che Liu and Junde Wu and Fenglin Liu and et al.},
      year={2025},
      eprint={2502.19634},
      archivePrefix={arXiv}
}

@misc{jiang2025alphadriveunleashingpowervlms,
      title={AlphaDrive: Unleashing the Power of VLMs in Autonomous Driving via Reinforcement Learning and Reasoning}, 
      author={Bo Jiang and Shaoyu Chen and Qian Zhang and Wenyu Liu and Xinggang Wang},
      year={2025},
      eprint={2503.07608},
      archivePrefix={arXiv}
}

@inproceedings{DBLP:conf/nips/Ouyang0JAWMZASR22,
  author       = {Long Ouyang and
                  Jeffrey Wu and
                  Xu Jiang and
                  Diogo Almeida and et al.},
  title        = {Training language models to follow instructions with human feedback},
  booktitle    = {Advances in Neural Information Processing Systems 35: Annual Conference
                  on Neural Information Processing Systems 2022, NeurIPS 2022, New Orleans,
                  USA},
  year         = {2022}
}

@misc{deepseekai2025deepseekr1incentivizingreasoningcapability,
      title={DeepSeek-R1: Incentivizing Reasoning Capability in LLMs via Reinforcement Learning}, 
      author={DeepSeek-AI},
      year={2025},
      eprint={2501.12948},
      archivePrefix={arXiv}
}

@misc{TheC3,
  author       = {Anthropic},
  title        = {The Claude 3 Model Family: Opus, Sonnet, Haiku},
  year         = {2024},
  howpublished = {\url{https://www.anthropic.com/news/claude-3-family}},
  note         = {Accessed: 2025-04-29}
}

@misc{geminiteam2024gemini15unlockingmultimodal,
      title={Gemini 1.5: Unlocking multimodal understanding across millions of tokens of context}, 
      author={Gemini Team},
      year={2024},
      eprint={2403.05530},
      archivePrefix={arXiv}
}

@misc{openai2024gpt4technicalreport,
      title={GPT-4 Technical Report}, 
      author={OpenAI},
      year={2024},
      eprint={2303.08774},
      archivePrefix={arXiv},
      primaryClass={cs.CL},
      url={https://arxiv.org/abs/2303.08774}, 
}

@misc{merchant2024generatingcontextuallyrelevantnavigationinstructions,
      title={Generating Contextually-Relevant Navigation Instructions for Blind and Low Vision People}, 
      author={Zain Merchant and Abrar Anwar and Emily Wang and Souti Chattopadhyay and Jesse Thomason},
      year={2024},
      eprint={2407.08219},
      archivePrefix={arXiv}
}

@inproceedings{DBLP:conf/cvpr/QiW0WWSH20,
  author       = {Yuankai Qi and
                  Qi Wu and
                  Peter Anderson and
                  Xin Wang and
                  William Yang Wang and
                  Chunhua Shen and
                  Anton van den Hengel},
  title        = {{REVERIE:} Remote Embodied Visual Referring Expression in Real Indoor
                  Environments},
  booktitle    = {2020 {IEEE/CVF} Conference on Computer Vision and Pattern Recognition,
                  {CVPR} 2020, Seattle, WA, USA},
  pages        = {9979--9988},
  publisher    = {Computer Vision Foundation / {IEEE}},
  year         = {2020}
}

@inproceedings{DBLP:conf/cvpr/AndersonWTB0S0G18,
  author       = {Peter Anderson and
                  Qi Wu and
                  Damien Teney and
                  Jake Bruce and
                  Mark Johnson and
                  Niko S{\"{u}}nderhauf and
                  Ian D. Reid and
                  Stephen Gould and
                  Anton van den Hengel},
  title        = {Vision-and-Language Navigation: Interpreting Visually-Grounded Navigation
                  Instructions in Real Environments},
  booktitle    = {2018 {IEEE} Conference on Computer Vision and Pattern Recognition,
                  {CVPR} 2018, Salt Lake City, UT, USA},
  pages        = {3674--3683},
  publisher    = {Computer Vision Foundation / {IEEE} Computer Society},
  year         = {2018}
}

@inproceedings{DBLP:conf/nips/Dai0LTZW0FH23,
  author       = {Wenliang Dai and
                  Junnan Li and
                  Dongxu Li and
                  Anthony Meng Huat Tiong and et al.},
  title        = {InstructBLIP: Towards General-purpose Vision-Language Models with
                  Instruction Tuning},
  booktitle    = {Proceceedings of the Annual Conference
                  on Neural Information Processing Systems 2023, NeurIPS 2023, New Orleans,
                  USA},
  year         = {2023}
}

@inproceedings{DBLP:conf/nips/LiuLWL23a,
  author       = {Haotian Liu and
                  Chunyuan Li and
                  Qingyang Wu and
                  Yong Jae Lee},
  title        = {Visual Instruction Tuning},
  booktitle    = {Advances in Neural Information Processing Systems 36: Annual Conference
                  on Neural Information Processing Systems 2023, NeurIPS 2023, New Orleans, USA},
  year         = {2023}
}

@misc{zhao2024vialmsurveybenchmarkvisually,
      title={VIALM: A Survey and Benchmark of Visually Impaired Assistance with Large Models}, 
      author={Yi Zhao and Yilin Zhang and Rong Xiang and Jing Li and Hillming Li},
      year={2024},
      eprint={2402.01735},
      archivePrefix={arXiv}
}

@misc{yuan2025walkvlmaidvisuallyimpairedpeople,
      title={WalkVLM:Aid Visually Impaired People Walking by Vision Language Model}, 
      author={Zhiqiang Yuan and Ting Zhang and Ying Deng and et al.},
      year={2025},
      eprint={2412.20903},
      archivePrefix={arXiv}
}

@inproceedings{DBLP:conf/acl/GopinathanMAS24,
  author       = {Muraleekrishna Gopinathan and
                  Martin Masek and
                  Jumana Abu{-}Khalaf and
                  David Suter},
  title        = {Spatially-Aware Speaker for Vision-and-Language Navigation Instruction
                  Generation},
  booktitle    = {Proceedings of the 62nd Annual Meeting of the Association for Computational
                  Linguistics (Volume 1: Long Papers), {ACL} 2024, Bangkok, Thailand},
  pages        = {13601--13614},
  publisher    = {Association for Computational Linguistics},
  year         = {2024}
}

@inproceedings{DBLP:conf/eccv/FanLWY24,
  author       = {Sheng Fan and
                  Rui Liu and
                  Wenguan Wang and
                  Yi Yang},
  title        = {Navigation Instruction Generation with {BEV} Perception and Large
                  Language Models},
  booktitle    = {Computer Vision - {ECCV} 2024 - 18th European Conference, Milan, Italy,
                  September 29-October 4, 2024, Proceedings, Part {XXII}},
  series       = {Lecture Notes in Computer Science},
  volume       = {15080},
  pages        = {368--387},
  publisher    = {Springer},
  year         = {2024}
}

@inproceedings{DBLP:conf/eccv/KongCWSHYL24,
  author       = {Xianghao Kong and
                  Jinyu Chen and
                  Wenguan Wang and
                  Hang Su and
                  Xiaolin Hu and
                  Yi Yang and
                  Si Liu},
  title        = {Controllable Navigation Instruction Generation with Chain of Thought
                  Prompting},
  booktitle    = {Computer Vision - {ECCV} 2024 - 18th European Conference, Milan, Italy,
                  September 29-October 4, 2024, Proceedings, Part {XXIX}},
  series       = {Lecture Notes in Computer Science},
  volume       = {15087},
  pages        = {37--54},
  publisher    = {Springer},
  year         = {2024}
}

@inproceedings{DBLP:conf/eccv/HuangSZBBO22,
  author       = {Zanming Huang and
                  Zhongkai Shangguan and
                  Jimuyang Zhang and et al.},
  title        = {{ASSISTER:} Assistive Navigation via Conditional Instruction Generation},
  booktitle    = {Computer Vision - {ECCV} 2022 - 17th European Conference, Tel Aviv,
                  Israel, October 23-27, 2022, Proceedings, Part {XXXVI}},
  series       = {Lecture Notes in Computer Science},
  volume       = {13696},
  pages        = {271--289},
  publisher    = {Springer},
  year         = {2022}
}

@article{giudice2008blind,
  title={Blind navigation and the role of technology},
  author={Giudice, Nicholas A and Legge, Gordon E},
  journal={The engineering handbook of smart technology for aging, disability, and independence},
  pages={479--500},
  year={2008},
  publisher={John Wiley \& Sons, Inc. Hoboken, NJ, USA}
}

@article{younis2019smart,
  title={A smart context-aware hazard attention system to help people with peripheral vision loss},
  author={Younis, Ola and Al-Nuaimy, Waleed and Rowe, Fiona and Alomari, Mohammad H},
  journal={Sensors},
  volume={19},
  number={7},
  pages={1630},
  year={2019},
  publisher={MDPI}
}
\end{document}